\documentclass[conference]{IEEEtran}
\IEEEoverridecommandlockouts
\usepackage{cite}
\usepackage{amsmath,amssymb,amsfonts}
\usepackage{algorithmic}
\usepackage{graphicx}
\usepackage{textcomp}
\usepackage[table]{xcolor}
\definecolor{lightgreen}{RGB}{144,238,144}

\usepackage{amsmath, amssymb}
\usepackage{graphicx}
\usepackage{amsmath}
\usepackage{amssymb}
\usepackage{booktabs}
\usepackage{comment}
\usepackage{xspace}
\usepackage{float} 
\usepackage{enumitem}

\usepackage{times}
\usepackage{cmap} 
\usepackage[T1]{fontenc} 
\usepackage{lmodern}
\usepackage{newtxtext,newtxmath}
\usepackage[a-1b]{pdfx}

\newcommand{\ourmodel}{\textbf{\textit{Point-LN}}\xspace}

\def\BibTeX{{\rm B\kern-.05em{\sc i\kern-.025em b}\kern-.08em
    T\kern-.1667em\lower.7ex\hbox{E}\kern-.125emX}}

\PassOptionsToPackage{pagebackref,breaklinks,colorlinks}{hyperref}
\usepackage{hyperref}
\hypersetup{
     colorlinks=true,       
     linkcolor=blue,        
     citecolor=red,         
     urlcolor=blue          
 }

\usepackage[capitalize]{cleveref}
\crefname{section}{Sec.}{Secs.}
\Crefname{section}{Section}{Sections}
\Crefname{table}{Table}{Tables}
\crefname{table}{Tab.}{Tabs.}

\IEEEpubid{\makebox[\columnwidth]{979-8-3315-2311-4/25/\$31.00~\copyright2025 IEEE \hfill} \hspace{\columnsep}\makebox[\columnwidth]{ }}

\makeatletter
\newcommand*\titleheader[1]{\gdef\@titleheader{#1}}
\AtBeginDocument{%
  \let\st@red@title\@title
  \def\@title{%
    \bgroup\normalfont\large\centering\@titleheader\par\egroup
    \vskip0.2em  
    \st@red@title}
}
\makeatother

\title{Point-LN: A Lightweight Framework for Efficient Point Cloud Classification Using Non-Parametric Positional Encoding}
\titleheader{2025 29th International Computer Conference, Computer Society of Iran (CSICC)}

\author{
\IEEEauthorblockN{Marzieh Mohammadi}
\IEEEauthorblockA{Sirjan University of Technology, Iran \\
\texttt{\small m.mohammadi@stu.sirjantech.ac.ir}}

\and
\IEEEauthorblockN{Amir Salarpour}
\IEEEauthorblockA{Sirjan University of Technology, Iran \\
\texttt{\small salarpour@sirjantech.ac.ir}}

\and
\IEEEauthorblockN{Pedram MohajerAnsari}
\IEEEauthorblockA{Clemson University, USA \\
\texttt{\small pmohaje@clemson.edu}}
}
\begin{document}

\maketitle
\IEEEpubidadjcol 

\begin{abstract}
We introduce \ourmodel, a novel lightweight framework engineered for efficient 3D point cloud classification. \ourmodel integrates essential non-parametric components—such as Farthest Point Sampling (FPS), k-Nearest Neighbors (k-NN), and non-learnable positional encoding—with a streamlined learnable classifier that significantly enhances classification accuracy while maintaining a minimal parameter footprint. This hybrid architecture ensures low computational costs and rapid inference speeds, making \ourmodel ideal for real-time and resource-constrained applications. Comprehensive evaluations on benchmark datasets, including ModelNet40 and ScanObjectNN, demonstrate that \ourmodel achieves competitive performance compared to state-of-the-art methods, all while offering exceptional efficiency. These results establish \ourmodel as a robust and scalable solution for diverse point cloud classification tasks, highlighting its potential for widespread adoption in various computer vision applications. For more details, see the code at: \url{https://github.com/asalarpour/Point_LN}.

\end{abstract}



\section{Introduction}
\label{sec:intro}
In the field of computer vision (CV), the technology for processing two-dimensional images nears maturity \cite{zhang2021joint} \cite{ning2022hcfnn}, and researchers are increasingly shifting their focus toward three-dimensional scenes, which more accurately represent real-world environments. Point clouds, consisting of large collections of tiny points in 3D space, play a crucial role in this process due to their rich informational content. These unordered sets of multidimensional points become meaningful when considered together, with point cloud classification being a key task in point cloud analysis. This task is essential across various domains, including security detection \cite{ning2021jwsaa, salarpour2014long} \cite{yan2021beyond}, target object detection \cite{zhangyu2021camera}, three-dimensional reconstruction \cite{yang2016automated}, and autonomous vehicles \cite{mohajeransari2024discovering, aldeenwip, 10588595}.

Point cloud data, due to its inherent irregularity and sparsity, presents significant challenges for processing and analysis. Deep learning-based models for point cloud classification have gained widespread adoption due to their advantages in feature extraction and high classification accuracy. Early methods such as PointNet \cite{qi2017pointnet} and PointNet++ \cite{qi2017pointnet++} introduced innovative frameworks for directly processing unordered point clouds, achieving remarkable performance. Subsequent approaches, including PointConv \cite{wu2019pointconv} and PointMLP \cite{ma2022rethinking}, further advanced the field by incorporating more sophisticated feature aggregation mechanisms, leading to state-of-the-art results in various benchmarks.

Despite their success, these parametric models have notable limitations. They often require large amounts of training data, substantial computational resources, and extensive hyperparameter tuning to achieve optimal performance. Moreover, their ability to generalize can be compromised in scenarios involving diverse datasets or varying conditions, limiting their adaptability in practice.

To address these challenges, non-parametric methods such as Point-NN \cite{zhang2023parameter} and Point-GN \cite{mohammadi2024point} have emerged as promising alternatives to parametric models in point cloud analysis. These methods eliminate the need for extensive parameter optimization and training, instead relying on techniques like nearest-neighbor search and similarity-based approaches. Their simplicity, flexibility, and ability to perform inference across diverse data distributions make them particularly suitable for scalable and robust solutions in tasks such as object detection and 3D reconstruction.

\begin{figure}[t]
    \centering
    \includegraphics[width=0.8\columnwidth]{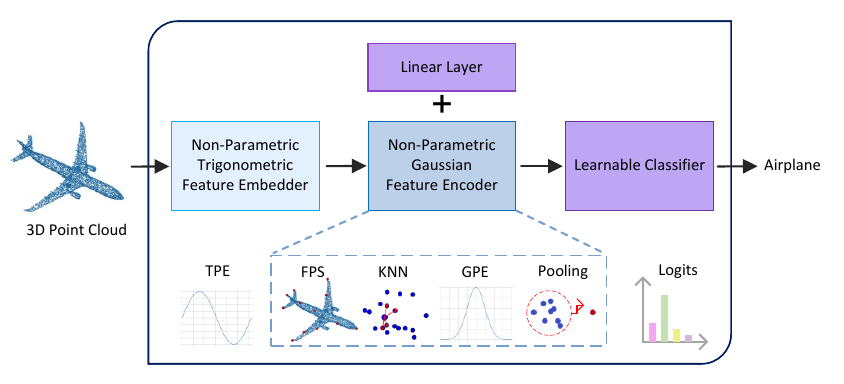}
    \vspace{-3mm}
    \caption{Illustration of \ourmodel for Point Cloud Classification.}
    \vspace{-5mm}
    \label{fig:illustration}
\end{figure}

However, despite their adaptability, non-parametric methods often struggle to achieve high accuracy in complex scenarios. To address these limitations, we propose a novel hybrid framework, \ourmodel, which combines the strengths of non-parametric techniques with a learnable classifier. This approach improves classification accuracy while maintaining reduced reliance on learnable parameters, as illustrated in \autoref{fig:illustration}. The proposed framework enhances versatility and effectiveness, making it highly applicable to real-world tasks like object detection and 3D reconstruction.


\begin{figure*}[ht] 
    \centering
    \includegraphics[width=0.65\textwidth]{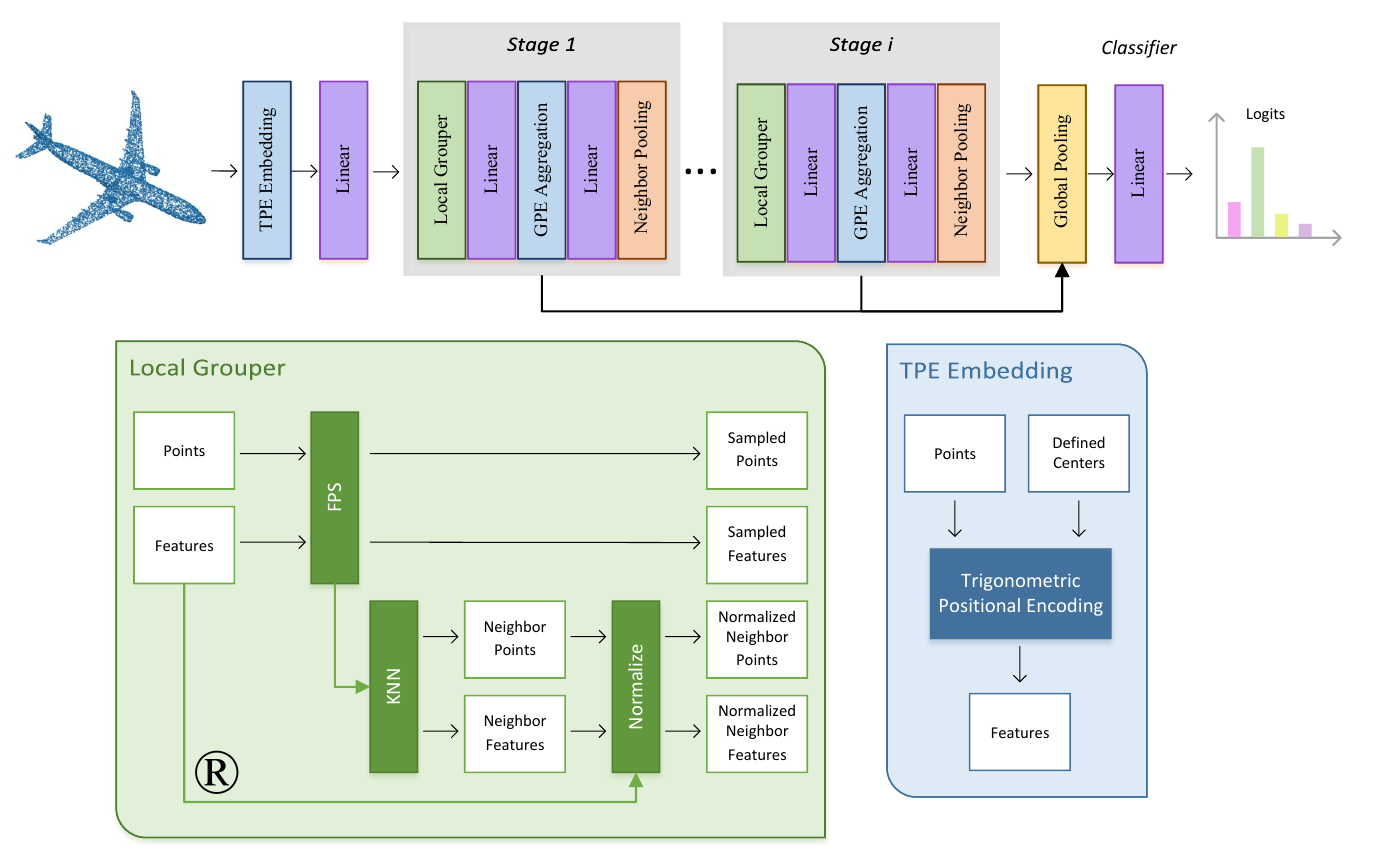}
    \vspace{-2mm}
    \caption{Overview of the proposed network architecture: the feature encoder extracts high-dimensional representations from raw point clouds, and the classifier maps these features to the target label space.}
    \label{fig:framework}
\end{figure*}

In this paper, we introduce an efficient network for point cloud classification that effectively captures and adaptively aggregates multivariate geometric features, without relying on learnable parameters for feature extraction. Additionally, we incorporate a classifier that uses a minimal number of learnable parameters, thereby enhancing both efficiency and performance.

The primary contributions of this paper are as follows:
\begin{itemize}
    \item We present a parameter-efficient derivative of non-parametric methods, delivering exceptional performance without relying on complicated operators.
    \item We incorporate a learnable classifier that effectively maps the extracted features to target categories, allowing for fine-tuning and improved classification performance.
    \item Despite its simple design philosophy, our approach demonstrates strong performance in 3D point cloud classification, achieving competitive results on popular benchmarks such as ModelNet40 \cite{wu20153d} and ScanObjectNN \cite{uy2019revisiting}.
\end{itemize}


\section{Related works}
\label{sec:related}
Methods for 3D point cloud classification can be broadly categorized into \textit{\textbf{Projection-based}} and \textit{\textbf{Point-based}} approaches. The following subsections review these two categories in detail and discuss their key techniques, advantages, and limitations.

\subsection{Projection-based methods}

Projection-based methods address the irregularity and sparsity of point clouds by transforming them into grid-like representations, such as multi-view depth maps \cite{bai2016gift} \cite{hamdi2021mvtn} and 3D voxels \cite{zhou2018voxelnet} \cite{li2016fpnn}. By utilizing sets of 2D views, which contain extensive information for recognizing 3D shapes, these methods, like MVCNN \cite{su2015multi}, use rendered 2D images to identify 3D objects. Similarly, VoxNet \cite{maturana2015voxnet} introduced a convolutional neural network (CNN) architecture that represents 3D data using a volumetric occupancy grid. While effective in processing 3D data, projection-based methods have limitations. Transforming point clouds into 2D or voxelized grids can result in the loss of fine-grained spatial relationships, potentially affecting classification performance, especially for complex tasks that require preserving precise geometric details.

\subsection{Point-based Methods}

Recent advances have focused on directly processing point clouds without transforming them into grids. \textit{Point-based methods} can be broadly categorized into \textit{\textbf{Parametric}} (deep learning-based) and \textit{\textbf{Non-parametric}} methods, as follows:

\subsubsection{\textbf{Parametric Methods}}

They typically rely on deep learning techniques and can be subdivided into several groups:

\begin{enumerate}[label=(\roman*)]

\item \textbf{Convolution-based methods} \cite{wu2019pointconv, li2018pointcnn, thomas2019kpconv, peyghambarzadeh2020point} use convolutional neural networks (CNN) to extract spatial features from point clouds. By applying convolutions over local neighborhoods of points, these methods capture spatial hierarchies and local geometric structures. This technique is highly effective for tasks such as classification, segmentation, and object recognition. For example, PointNet \cite{qi2017pointnet} and PointNet++ \cite{qi2017pointnet++} employ convolutional layers to learn point-wise features, aggregating them to form a global representation of the point cloud. These models excel at handling unordered, irregular point clouds, making them suitable for a variety of 3D vision tasks.

\item  \textbf{Graph-based methods} \cite{lu2020pointngcnn, wang2019dynamic}  treat point clouds as graphs, where each point is represented as a node, and edges denote relationships or distances between points. This representation is particularly effective for capturing connectivity and local structures within the point cloud. Notably, Dynamic Graph CNN (DGCNN) \cite{wang2019dynamic} dynamically constructs graphs during training, allowing the model to learn both local and global features. This dynamic graph construction improves the model’s ability to capture complex geometric relationships, enhancing performance on tasks like classification and segmentation.

\item \textbf{Transformer-based methods} \cite{yang2019modeling} \cite{engel2021point} leverage self-attention mechanisms to process point clouds, allowing the model to weigh the importance of different points based on their spatial relationships. This ability to capture long-range dependencies is particularly useful for tasks where global context is crucial. For example, Point Transformer \cite{engel2021point} introduces ScorNet, a learning score-based focus module. 

Point Transformer first extracts local and global features and then uses ScorNet to rank these local features. A local-global attention mechanism is applied to associate the ranked local features with global features, allowing the model to focus on the most relevant features for classification and segmentation. While parametric models have demonstrated high accuracy, they often come with high computational complexity, limiting their applicability in real-time scenarios.

\end{enumerate}

\subsubsection{\textbf{Non-parametric Methods}} These methods, unlike parametric ones, do not require training or parameter tuning, which eliminates much of the computational overhead. These methods focus on directly capturing geometric features from point clouds, making them computationally efficient and adaptable to diverse data without the need for extensive training.

\begin{enumerate}[label=(\roman*)]

\item \textbf{Point-NN} \cite{zhang2023parameter} leverages non-parametric techniques to effectively capture geometric features by focusing on local neighborhood information. By minimizing reliance on learnable parameters, Point-NN enhances adaptability while maintaining simplicity. 

\item \textbf{Point-GN} \cite{mohammadi2024point} Similarly builds on these concepts by introducing a Gaussian embedding function that improves classification accuracy without adding significant computational overhead. Despite their simplicity and flexibility, non-parametric methods often fall short of the high accuracy levels seen in parametric models, especially for complex classification tasks. This trade-off between simplicity and performance underscores the need for further advancements in the field.

\item \textbf{Point-PN} \cite{zhang2023parameter} effectively combines the strengths of both parametric and non-parametric methods to bridge this gap. By integrating a conventional learnable classifier and enhancing the embedding with parametric linear layers, Point-PN captures higher-level spatial patterns while maintaining simplicity. Additionally, it employs trigonometric positional encoding to better represent geometric relationships within the point cloud, enabling the model to extract rich spatial features. This approach demonstrates that a powerful and efficient 3D model can be developed from a non-parametric framework without relying on complex operators or an excessive number of parameters.

\end{enumerate}

\section{Method}
\label{sec:method}
In this section, we first introduce the fundamentals of 3D point clouds and review common classification techniques. Next, we explore the application of non-parameteric positional encodings in lightweight methods and how they enhance the processing of point cloud data. We then outline the architecture of our proposed network \ourmodel, which consists of two key components: the \textit{\textbf{feature encoder}} and the \textit{\textbf{classifier}}. The feature encoder transforms the raw input point cloud into high-dimensional feature representations, while the classifier maps these encoded features to the target label space, producing classification logits. A detailed overview of the architecture is provided in~\autoref{fig:framework}.


\subsection{Background}
Point cloud classification aims to assign a label to an entire point cloud, or to specific regions within the point cloud, allowing for the identification of object characteristics. A 3D point cloud is a set of points $\mathcal{P} = \{p_1, p_2, \dots, p_N\}$, where each point $p_i$ is represented by its 3D coordinates $(x_i, y_i, z_i)$, and optionally additional features $\gamma(p_i) \in \mathbb{R}^{C \times 3}$, where $C$ is the feature dimension. The goal is to process this input point cloud and classify it into one of several predefined categories. The classification process consists of two main steps: the feature encoder and the classifier. First, the feature encoder extracts meaningful representations from the raw point cloud by aggregating spatial and geometric information across points. The encoded feature vector $\mathbf{F}$ represents the output of the feature encoder, which processes the input point cloud:

\begin{equation}
\mathbf{F} = \text{Encoder}(\{p_i\}_{i=1}^N)
\end{equation}

The second step, the classifier, takes $\mathbf{F}$ as input to predict the point cloud's category, typically using fully connected layers followed by a softmax activation. The classification output $\mathbf{y} $ is given by:
\begin{equation}
\mathbf{y}  = \text{Softmax}(\text{Classifier}(\mathbf{F}))
\end{equation}


\subsection{Positional Encoding}
Positional encoding (PE), first introduced in the transformer architecture \cite{vaswani2017attention}, has proven successful in various domains such as natural language processing (NLP) and CV. For point cloud processing, positional encoding captures essential spatial relationships between points, enabling the model to understand their relative positions in 3D space. We explore two types of positional encoding commonly used in point cloud processing.


\begin{enumerate}[label=(\roman*)]
\item \textbf{Trigonometric Positional Encoding (TPE)}

Point-NN \cite{zhang2023parameter} employs TPE, taking advantage of the periodic nature of sine and cosine functions. This method effectively captures the spatial relationships between points in 3D space. For a point $p_i = (x_i, y_i, z_i)$, trigonometric functions are used to generate a $C_I$-dimensional positional embedding. The encoding for each axis is defined as follows:

\begin{equation}
\text{PE}_{\text{TPE}}\quad
\begin{cases} 
\gamma_x(x_i,2n) = \sin(\alpha x_i /\beta ^{\frac{6n}{C_I}}) \\
\gamma_x(x_i,2n + 1) = \cos(\alpha x_i /\beta ^{\frac{6n}{C_I}}) \\
\gamma_y(y_i,2n) = \sin(\alpha y_i /\beta ^{\frac{6n}{C_I}}) \\
\gamma_y(y_i,2n + 1) = \cos(\alpha y_i /\beta ^{\frac{6n}{C_I}}) \\
\gamma_z(z_i,2n) = \sin(\alpha z_i /\beta ^{\frac{6n}{C_I}}) \\
\gamma_z(z_i,2n + 1) = \cos(\alpha z_i /\beta ^{\frac{6n}{C_I}}) \\
\end{cases}
\end{equation}

\noindent where $C_I$ is the initial feature dimension, $\alpha$ controls the scale, and $\beta$ governs the wavelength. The final positional encoding for each point is given by:


{\small
\begin{equation}
\gamma^{TPE}(p_i) = \text{Concat} (\gamma_x({x_i}); \gamma_y({y_i}); \gamma_z({z_i})) \in \mathbb{R}^{1 \times C_I}
\end{equation}
}




\item \textbf{Gaussian Positional Encoding (GPE)}
GPE, used in Point-GN \cite{mohammadi2024point}, enhances spatial awareness by encoding positional information into the feature representation without requiring learnable parameters. GPE uses a Gaussian function to map the raw point coordinates into higher-dimensional space, preserving both local and global spatial information. For each axis, the encoding is defined as:


\begin{equation}
\text{PE}_{\text{GPE}} \quad
\begin{cases}
\gamma_x(x_i, v_j) = \exp\left( -\frac{\| x_i^2 - v_j \|^2}{2\sigma^2} \right) \\
\gamma_y(y_i, v_j) = \exp\left( -\frac{\| y_i^2 - v_j \|^2}{2\sigma^2} \right) \\
\gamma_z(z_i, v_j) = \exp\left( -\frac{\| z_i^2 - v_j \|^2}{2\sigma^2} \right)
\end{cases}
\end{equation}

\noindent where $v_j$ represents reference points and $\sigma$ is the standard deviation that influences the scale of local versus global features. The final positional encoding for each point is the concatenation of embeddings across all axes:

\begin{equation}
\gamma^{GPE}(p_i) = \left[\gamma_x(x_i, v_j), \gamma_y(y_i, v_j), \gamma_z(z_i, v_j)\right]_{j=1}^V
\end{equation}

\noindent where \( V \) is the number of reference points along each axis.

\end{enumerate}


\subsection{Lightweight Feature Encoder}
We provide a lightweight feature encoder, and to better extract multi-scale hierarchy, we append simple linear layers to each stage of the encoder. In the following sections, we will outline the four steps of this feature encoder.


\begin{enumerate}

\vspace{2mm}
\item \textit{Initial Embedding}
\vspace{1mm}

\textbf{TPE.} To perform feature embedding  we utilize \textit{TPE}, which transforms the Cartesian coordinates of each point \(p_i \in \mathbb{R}^d\) into a trigonometric representation, preserving the spatial relationships and helping the model maintain a consistent understanding of geometry. The resulting embeddings, \(\gamma^{TPE}(p_i)\), encapsulate geometric features derived from the point cloud.

\textbf{Linear Layer.} The trigonometric embeddings \(\gamma^{TPE}(p_i)\) are then passed through a linear layer, which performs a learnable linear transformation. This operation is defined as:  
\[
f_i = \text{Linear}(\gamma^{TPE}(p_i)) = \mathbf{W} \cdot \gamma^{TPE}(p_i) + \mathbf{b},
\]  
where \(\mathbf{W}\) is a learnable weight matrix, \(\mathbf{b}\) is a bias vector, and \(f_i\) represents the output embedded features for point \(p_i\). This step embeds the TPE-transformed features, enabling the model to capture more complex patterns in the input data.

\textbf{Final Output.}  The output of the initial embedding process includes the original points and their associated embedded features, represented as:  
\[
\{p_i, f_i\}_{i=1}^N,
\]  
where \(p_i\) are the original Cartesian coordinates of the points, and \(f_i = \text{Linear}(\gamma^{TPE}(p_i))\) are the embedded features derived from the TPE-transformed coordinates.


\vspace{2mm}
\item \textit{Local Grouper}
\vspace{1mm}

\textbf{Sampling Layer.} At each stage, the input point cloud and associated features from the previous stage is represented as
\(\{p_i, f_i\}_{i=1}^N\) the features could be TPE from intial embed step or GPE for different stages. To reduce computational complexity while retaining the global structure of the point cloud, we perform \textit{Farthest Point Sampling (FPS)}. This step selects a subset of $N/2$ points that are maximally spaced apart, ensuring they represent the center of the point cloud:

\begin{equation}
\{\mathbf{p}_j, f_j\}_{j=1}^{N/2} = \text{FPS}\left(\{\mathbf{p}_i, f_i\}_{i=1}^N\right)
\end{equation}

\textbf{Grouping Layer.} We then group the points into local neighborhoods using the \textit{K-Nearest Neighbors (KNN)} algorithm. This step allows us to capture local geometric patterns:

\begin{equation}
\text{idx}_j = \text{KNN}(\mathbf{p}_j, \mathbf{p}_i)
\end{equation}

\noindent  where $\text{idx}_j$ represents the indices of the $K$ nearest neighbors for point $\mathbf{p}_j$.
The retrieved coordinates and features are:
\begin{equation}
\mathbf{P}_j = \text{retrieve}\left(\{\mathbf{p}_i\}_{i=1}^N ,\text{idx}_j\right) \in \mathbb{R}^{K \times 3}
\end{equation}
\begin{equation}
\mathbf{F}_j = \text{retrieve}\left(\{f_i)\}_{i=1}^N ,\text{idx}_j\right) \in \mathbb{R}^{K \times (V \times 3)}
\end{equation}
where, $\mathbf{P}_j$ denotes the gathered coordinates, and $\mathbf{F}_j$ represents the gathered features for the point $\mathbf{p}_j$. The retrieved coordinates $\mathbf{P}_j$ and features $\mathbf{F}_j$ are then normalized using the mean and standard deviation of each point’s neighbors. These normalized coordinates and features are subsequently forwarded to the next stage for further processing.

\vspace{2mm}
\item \textit{Local Geometry Aggregation}
\vspace{1mm}

\textbf{GPE Aggregation.} The features from the Local Grouper are then fed into the
\textit{Local Geometry Aggregation}, Here, we use \textit{GPE} to extract meaningful spatial information, that helps encode global and local geometric patterns, ensuring a rich feature representation without the need for learnable parameters.
The spatially encoded features are combined with the neighborhood features through element-wise multiplication:

\begin{equation}
\mathbf{F}_j \leftarrow \mathbf{F}_j + \gamma(\mathbf{P}_j) \odot \gamma(\mathbf{P}_j)
\end{equation}
\textbf{Linear Layer.} Here, we insert two learnable linear layers for each stage, positioned right before and after the GPE Aggregation step. This addition aims to capture higher-level spatial information. Specifically, by placing these layers strategically, we enhance the encoder's ability to process and represent geometric features effectively.

\vspace{2mm}
\item \textit{Pooling}
\vspace{1mm}

\textbf{Neighbor Pooling.} To aggregate local features, we use both  mean and max pooling, ensuring that the final features are permutation-invariant and capture local geometry from multiple perspectives:
\begin{equation}
\Phi_j = \text{Mean}(\mathbf{F}_j) + \text{Max}(\mathbf{F}_j), \, \forall j \in \mathbb{Z}_K
\end{equation}

\noindent  where \( \text{Mean}(\mathbf{F}_j) \) and \( \text{Max}(\mathbf{F}_j) \) are permutation-invariant operations. These operations capture neighboring features from multiple perspectives and ensure that the order of neighbors has no effect on the final pooled features.

\textbf{Aggregation Across Stages.} The Lightweight feature encoder includes four stages, each producing pooled features $\Phi_j^s$. After processing through all stages, global pooling is performed on the results from each stage. The final feature vector $\mathbf{F}$ for the lightweight feature encoder is obtained by concatenating the global mean and max features from all four stages:

\begin{equation} 
\mathbf{F} = \bigoplus_{s=1}^4 \left[\text{Mean}(\Phi_j^s) + \text{Max}(\Phi_j^s)\right] 
\end{equation}
which captures and aggregates spatial and feature information across multiple levels by integrating the mean and max features from each stage.

\end{enumerate}


\subsection{Classifier Architecture}

The classifier maps the high-dimensional feature space, produced by the lightweight encoder, directly to the target label space. This mapping is performed by a simple neural network, which generates the output logits. The predicted probabilities are then obtained using the softmax function:

\begin{equation}
\mathbf{y} = \text{Softmax}(\text{Classifier}(\mathbf{F})),
\end{equation}

where \(\text{Classifier}(\cdot)\) represents the transformation from the feature space \(\mathbf{F}\) to the logits.

The final predicted class is determined by selecting the class with the highest probability:

\begin{equation}
c = \arg \max ({\mathbf{y}}).
\end{equation}








\section{Experiments}
\label{sec:exper}
In this section, we evaluate the performance of our method for point cloud classification through experiments conducted on two well-known datasets: ModelNet40 \cite{wu20153d} and ScanObjectNN \cite{uy2019revisiting}. ModelNet40 consists of clean, synthetic 3D models, providing a controlled environment for assessing classification performance on ideal data. In contrast, ScanObjectNN presents a more challenging scenario with real-world 3D data that includes occlusions, clutter, and background noise, thus offering a more rigorous test of the robustness and versatility of our approach.

To thoroughly evaluate our method, we compare it against several state-of-the-art techniques, including both fully trained models and non-parametric methods. Specifically, we benchmark our approach against PointMLP \cite{ma2022rethinking}, a recent fully trained model that has demonstrated impressive results in point cloud classification tasks. Additionally, we assess the performance of non-parametric methods such as Point-NN \cite{zhang2023parameter} and Point-GN \cite{mohammadi2024point}, which have shown promising results in similar classification settings. Through these comprehensive comparisons, we aim to highlight the strengths of our method, demonstrating its effectiveness in handling both synthetic and real-world data, while also emphasizing its competitive edge in the context of fully trained and non-parametric models.

\subsection{Experimental Setup}

We evaluate the performance of \ourmodel on a system equipped with an NVIDIA RTX 4090 GPU. Although \ourmodel is designed to be lightweight, the use of the high-performance GPU significantly accelerates both the training and inference processes. This allows for efficient model evaluation and rapid experimentation, particularly when dealing with large-scale and complex 3D datasets such as \textit{ModelNet40} \cite{wu20153d} and \textit{ScanObjectNN} \cite{uy2019revisiting}.

While \ourmodel does not require a vast amount of training resources, the GPU's computational power enables faster convergence and efficient handling of the diverse real-world data. This ensures that we can conduct comprehensive benchmarking, rapidly assess the model's performance, and explore various hyperparameter configurations. The use of this hardware ensures our method can be evaluated effectively across both synthetic and real-world datasets, highlighting its strengths in terms of efficiency and scalability.

\begin{table}[h]
\caption{Comparison of Point Cloud Classification Methods On ModelNet40}
\begin{center}
\begin{tabular}{l|cc}
\hline
\textbf{Method} & \textbf{Acc. (\%)} & \textbf{Param.} \\ 
\hline
Point-NN \cite{zhang2023parameter}   & 81.8  & 0.0 M \\ 
\hline
Point-GN  \cite{mohammadi2024point}  & 85.3  & 0.0 M  \\ 
\hline
PointNet \cite{qi2017pointnet}  & 89.2  & 3.5 M \\ 
\hline
PointNet++ \cite{qi2017pointnet++}  & 90.7  & 1.7 M \\ 
\hline
DGCNN \cite{wang2019dynamic}   & 92.9  & 1.8 M  \\ 
\hline
GBNet   \cite{qiu2021geometric}    & 93.8  & 8.4 M \\ 
\hline
CurveNet  \cite{xiang2021walk}  & 93.8  & 2.0 M  \\ 
\hline
Point-PN \cite{zhang2023parameter} & 93.8 & \textbf{0.8 M} \\
\hline
PointMLP \cite{ma2022rethinking}   & \textbf{94.1}  & 12.6 M \\ 
\hline

\rowcolor{green!15}  \textbf{\ourmodel}  & \textbf{94.0}   & \textbf{0.8 M} \\ 
\hline
\end{tabular}
\label{tab:modelnet40}
\end{center}
\vspace{-5mm}
\end{table}

\subsection{Dataset Details}

The \textit{ModelNet40} \cite{wu20153d} dataset consists of 12,311 CAD models across 40 object categories, split into 9,843 samples for training and 2,468 for testing. This dataset is widely used for point cloud classification due to its clean, synthetic nature, providing a controlled environment for benchmarking.

In contrast, the \textit{ScanObjectNN} \cite{uy2019revisiting} dataset presents a more challenging real-world scenario, with 2,902 samples across 15 object categories. Objects in ScanObjectNN are often occluded, cluttered, or contain background noise, providing a closer simulation to real-world 3D data. The dataset is divided into three official subsets: \textit{OBJ-BG}, which contains objects with background noise, \textit{OBJ-ONLY}, with objects without background, and \textit{PB-T50-RS}, featuring partial occlusions and transformations. These subsets test the robustness of models under various degrees of complexity. For both datasets, we follow the common practice of sampling 1,024 points from each object, as used in prior works (e.g., PointNet++ \cite{qi2017pointnet++}, DGCNN \cite{wang2019dynamic}). Our model combines \textit{maximum pooling} and \textit{average pooling} to enhance feature aggregation, inspired by DGCNN \cite{wang2019dynamic}.

\subsection{Shape classification on ModelNet40}
\autoref{tab:modelnet40} evaluates the performance of \ourmodel on the \textit{ModelNet40}~\cite{wu20153d} dataset. It achieves an accuracy of \textit{94.0\%}, which is comparable to the best performing model, PointMLP \cite{ma2022rethinking}, with an accuracy of 94.1\%. However, PointMLP has a significantly larger model size (12.6M parameters). In contrast, our approach demonstrates the ability to capture both local and global geometric features while maintaining minimal model complexity, with only \textit{0.8 million parameters}. When compared to other methods, \ourmodel shows competitive performance with models such as PointNet \cite{qi2017pointnet} (89.2\%) and PointNet++ \cite{qi2017pointnet++} (90.7\%), while being much more lightweight. 

For instance, \ourmodel achieves a higher accuracy than Point-GN \cite{mohammadi2024point} (85.3\%) and Point-NN \cite{zhang2023parameter} (81.8\%), both of which have zero parameters but show lower classification accuracy. This demonstrates the effectiveness of \ourmodel in extracting meaningful features with only a small number of parameters. Additionally, \ourmodel ensures high efficiency for real-time applications. Despite its compact size, it provides performance on par with larger parametric models like PointMLP, but with substantially reduced computational overhead. This combination of competitive accuracy and exceptional efficiency makes \ourmodel an ideal choice for resource-constrained environments, where both real-time performance and minimal model complexity are crucial.

\subsection{Shape classification on ScanObjectNN}
While the ModelNet40 \cite{wu20153d} dataset is a well-established benchmark for point cloud analysis, its synthetic nature may limit the evaluation of methods under more challenging, real-world conditions. To address this, we also, we evaluate the performance of \ourmodel on the challenging ScanObjectNN \cite{uy2019revisiting} benchmark, presenting our results in \autoref{tab:scan}. \ourmodel achieves state-of-the-art accuracy across all subsets, with \textit{92.2\%} on OBJ-BG, \textit{92.1\%} on OBJ-ONLY, and \textit{91.7\%} on the most challenging PB-T50-RS subset. These results highlight the robustness of \ourmodel in handling real-world 3D data with noise, occlusions, and clutter.

\begin{table}[h]
\caption{Comparison of Point Cloud Classification Methods On ScanObjectNN}
\begin{center}
\begin{tabular}{l|cccc}  
\hline
\textbf{Method} & \textbf{OBJ-BG} & \textbf{OBJ-ONLY} & \textbf{PB-T50-RS} & \textbf{Param.} \\ 

\hline
Point-NN \cite{zhang2023parameter} & 71.1  & 74.9  & 64.9  & 0.0 M \\ 
\hline
Point-GN \cite{mohammadi2024point} & 85.2  & 86.0  & 86.4  & 0.0 M \\ 
\hline
PointNet \cite{qi2017pointnet} & 73.3  & 79.2  & 68.2  & 3.5 M \\ 
\hline
PointNet++ \cite{qi2017pointnet++} & 82.3  & 84.3  & 77.9  & 1.7 M \\ 
\hline
DGCNN  \cite{wang2019dynamic}   & 82.8  & 86.2  & 78.1  & 1.8 M \\ 
\hline
PointCNN \cite{li2018pointcnn}  & 86.1  & 85.5  & 78.5  & -     \\ 
\hline
GBNet  \cite{qiu2021geometric}  & -     & -     & 80.5  & 8.4 M \\ 
\hline
PointMLP \cite{ma2022rethinking} & -     & -     & 85.4  & 12.6 M \\ 
\hline
Point-PN \cite{zhang2023parameter} & 91.0 & 90.2 & 87.1 & \textbf{0.8 M} \\

\hline
\rowcolor{green!15} \textbf{\ourmodel} & \textbf{92.2} & \textbf{92.1 } & \textbf{91.7}  &  \textbf{0.8 M} \\ 
\hline
\end{tabular}
\label{tab:scan}
\end{center}
\end{table}

Compared to parametric methods, \ourmodel delivers substantial accuracy improvements while maintaining a lightweight architecture. For instance, it outperforms PointNet \cite{qi2017pointnet} by \textit{+18.9\%} on OBJ-BG and PointNet++ \cite{qi2017pointnet++} by \textit{+9.9\%} on OBJ-ONLY. Furthermore, \ourmodel matches the performance of the larger PointMLP \cite{ma2022rethinking} (91.7\% vs. 85.4\% on PB-T50-RS) while requiring only \textit{0.8 million parameters}, compared to PointMLP’s \textit{12.6 million}. Similar to non-parametric methods such as Point-NN \cite{zhang2023parameter} and Point-GN \cite{mohammadi2024point}, \ourmodel leverages positional encoding for feature extraction without relying heavily on trainable parameters. 

However, \ourmodel introduces a novel lightweight architecture that significantly enhances performance, achieving a \textit{+10.3\%} improvement over Point-GN on PB-T50-RS while maintaining a small parameter count. These results demonstrate the effectiveness of \ourmodel in combining the strengths of non-parametric techniques with enhanced feature extraction capabilities. Its ability to deliver high accuracy with minimal computational overhead makes \ourmodel highly suitable for resource-constrained environments and real-time applications.

\section{Conclusion}

In this paper, we presented \textbf{\ourmodel}, a lightweight framework tailored for efficient 3D point cloud classification. By integrating essential non-parametric components—such as Farthest Point Sampling (FPS), k-Nearest Neighbors (k-NN), and non-learnable positional encoding—with a streamlined learnable classifier, \ourmodel achieves high classification accuracy with minimal parameters. This design ensures both computational efficiency and rapid inference, making it suitable for real-time and resource-constrained applications. Our experiments on benchmark datasets, ModelNet40 and ScanObjectNN, demonstrate that \ourmodel matches or outperforms state-of-the-art methods while maintaining a significantly smaller model size. 

The framework's robustness and adaptability are evident in its ability to handle diverse and noisy real-world 3D data effectively. Future work will explore enhancing \ourmodel by incorporating additional geometric features and extending its application to more complex 3D tasks such as semantic segmentation and object detection. These advancements aim to further improve scalability and performance, solidifying \ourmodel's role as a versatile tool in 3D point cloud analysis. In summary, \ourmodel offers a balanced solution that combines high accuracy with exceptional efficiency, positioning it as a promising framework for various computer vision applications involving 3D point clouds.



{\small
\bibliographystyle{IEEEtran}
\bibliography{mylib}

\begin{thebibliography}{10}
\providecommand{\url}[1]{#1}
\csname url@samestyle\endcsname
\providecommand{\newblock}{\relax}
\providecommand{\bibinfo}[2]{#2}
\providecommand{\BIBentrySTDinterwordspacing}{\spaceskip=0pt\relax}
\providecommand{\BIBentryALTinterwordstretchfactor}{4}
\providecommand{\BIBentryALTinterwordspacing}{\spaceskip=\fontdimen2\font plus
\BIBentryALTinterwordstretchfactor\fontdimen3\font minus \fontdimen4\font\relax}
\providecommand{\BIBforeignlanguage}[2]{{%
\expandafter\ifx\csname l@#1\endcsname\relax
\typeout{** WARNING: IEEEtran.bst: No hyphenation pattern has been}%
\typeout{** loaded for the language `#1'. Using the pattern for}%
\typeout{** the default language instead.}%
\else
\language=\csname l@#1\endcsname
\fi
#2}}
\providecommand{\BIBdecl}{\relax}
\BIBdecl

\bibitem{zhang2021joint}
L.~Zhang, L.~Sun, W.~Li, J.~Zhang, W.~Cai, C.~Cheng, and X.~Ning, ``A joint bayesian framework based on partial least squares discriminant analysis for finger vein recognition,'' \emph{IEEE Sensors Journal}, vol.~22, no.~1, pp. 785--794, 2021.

\bibitem{ning2022hcfnn}
X.~Ning, W.~Tian, Z.~Yu, W.~Li, X.~Bai, and Y.~Wang, ``Hcfnn: high-order coverage function neural network for image classification,'' \emph{Pattern Recognition}, vol. 131, p. 108873, 2022.

\bibitem{ning2021jwsaa}
X.~Ning, K.~Gong, W.~Li, and L.~Zhang, ``Jwsaa: joint weak saliency and attention aware for person re-identification,'' \emph{Neurocomputing}, vol. 453, pp. 801--811, 2021.

\bibitem{salarpour2014long}
A.~Salarpour, H.~Khotanlou, and N.~Mavridis, ``Long-term estimation of human spatial interactions through multiple laser ranging sensors,'' in \emph{2014 International Conference on Robotics and Emerging Allied Technologies in Engineering (iCREATE)}.\hskip 1em plus 0.5em minus 0.4em\relax IEEE, 2014, pp. 109--114.

\bibitem{yan2021beyond}
C.~Yan, G.~Pang, X.~Bai, C.~Liu, X.~Ning, L.~Gu, and J.~Zhou, ``Beyond triplet loss: person re-identification with fine-grained difference-aware pairwise loss,'' \emph{IEEE Transactions on Multimedia}, vol.~24, pp. 1665--1677, 2021.

\bibitem{zhangyu2021camera}
W.~Zhangyu, Y.~Guizhen, W.~Xinkai, L.~Haoran, and L.~Da, ``A camera and lidar data fusion method for railway object detection,'' \emph{IEEE Sensors Journal}, vol.~21, no.~12, pp. 13\,442--13\,454, 2021.

\bibitem{yang2016automated}
B.~Yang, R.~Huang, J.~Li, M.~Tian, W.~Dai, and R.~Zhong, ``Automated reconstruction of building lods from airborne lidar point clouds using an improved morphological scale space,'' \emph{Remote Sensing}, vol.~9, no.~1, p.~14, 2016.

\bibitem{mohajeransari2024discovering}
P.~MohajerAnsari, A.~Domeke, J.~de~Voor, A.~Mitra, G.~Johnson, A.~Salarpour, H.~Olufowobi, M.~Hamad, and M.~D. Pes{\'e}, ``Discovering new shadow patterns for black-box attacks on lane detection of autonomous vehicles,'' \emph{arXiv preprint arXiv:2409.18248}, 2024.

\bibitem{aldeenwip}
M.~Aldeen, P.~MohajerAnsari, J.~Ma, M.~Chowdhury, L.~Cheng, and M.~D. Pes{\'e}, ``Wip: A first look at employing large multimodal models against autonomous vehicle attacks,'' in \emph{ISOC Symposium on Vehicle Security and Privacy (VehicleSec '24)}, 2024.

\bibitem{10588595}
M.~Aldeen, P.~MohajerAnsari, J.~Ma, M.~Chowdhury, L.~Cheng, and M.~D. Pesé, ``An initial exploration of employing large multimodal models in defending against autonomous vehicles attacks,'' in \emph{2024 IEEE Intelligent Vehicles Symposium (IV)}, 2024, pp. 3334--3341.

\bibitem{qi2017pointnet}
C.~R. Qi, H.~Su, K.~Mo, and L.~J. Guibas, ``Pointnet: Deep learning on point sets for 3d classification and segmentation,'' in \emph{Proceedings of the IEEE conference on computer vision and pattern recognition}, 2017, pp. 652--660.

\bibitem{qi2017pointnet++}
C.~R. Qi, L.~Yi, H.~Su, and L.~J. Guibas, ``Pointnet++: Deep hierarchical feature learning on point sets in a metric space,'' \emph{Advances in neural information processing systems}, vol.~30, 2017.

\bibitem{wu2019pointconv}
W.~Wu, Z.~Qi, and L.~Fuxin, ``Pointconv: Deep convolutional networks on 3d point clouds,'' in \emph{Proceedings of the IEEE/CVF Conference on computer vision and pattern recognition}, 2019, pp. 9621--9630.

\bibitem{ma2022rethinking}
X.~Ma, C.~Qin, H.~You, H.~Ran, and Y.~Fu, ``Rethinking network design and local geometry in point cloud: A simple residual mlp framework,'' \emph{arXiv preprint arXiv:2202.07123}, 2022.

\bibitem{zhang2023parameter}
R.~Zhang, L.~Wang, Z.~Guo, Y.~Wang, P.~Gao, H.~Li, and J.~Shi, ``Parameter is not all you need: Starting from non-parametric networks for 3d point cloud analysis,'' \emph{arXiv preprint arXiv:2303.08134}, 2023.

\bibitem{mohammadi2024point}
M.~Mohammadi and A.~Salarpour, ``Point-gn: A non-parametric network using gaussian positional encoding for point cloud classification,'' \emph{arXiv preprint arXiv:2412.03056}, 2024.

\bibitem{wu20153d}
Z.~Wu, S.~Song, A.~Khosla, F.~Yu, L.~Zhang, X.~Tang, and J.~Xiao, ``3d shapenets: A deep representation for volumetric shapes,'' in \emph{Proceedings of the IEEE conference on computer vision and pattern recognition}, 2015, pp. 1912--1920.

\bibitem{uy2019revisiting}
M.~A. Uy, Q.-H. Pham, B.-S. Hua, T.~Nguyen, and S.-K. Yeung, ``Revisiting point cloud classification: A new benchmark dataset and classification model on real-world data,'' in \emph{Proceedings of the IEEE/CVF international conference on computer vision}, 2019, pp. 1588--1597.

\bibitem{bai2016gift}
S.~Bai, X.~Bai, Z.~Zhou, Z.~Zhang, and L.~Jan~Latecki, ``Gift: A real-time and scalable 3d shape search engine,'' in \emph{Proceedings of the IEEE conference on computer vision and pattern recognition}, 2016, pp. 5023--5032.

\bibitem{hamdi2021mvtn}
A.~Hamdi, S.~Giancola, and B.~Ghanem, ``Mvtn: Multi-view transformation network for 3d shape recognition,'' in \emph{Proceedings of the IEEE/CVF International Conference on Computer Vision}, 2021, pp. 1--11.

\bibitem{zhou2018voxelnet}
Y.~Zhou and O.~Tuzel, ``Voxelnet: End-to-end learning for point cloud based 3d object detection,'' in \emph{Proceedings of the IEEE conference on computer vision and pattern recognition}, 2018, pp. 4490--4499.

\bibitem{li2016fpnn}
Y.~Li, S.~Pirk, H.~Su, C.~R. Qi, and L.~J. Guibas, ``Fpnn: Field probing neural networks for 3d data,'' \emph{Advances in neural information processing systems}, vol.~29, 2016.

\bibitem{su2015multi}
H.~Su, S.~Maji, E.~Kalogerakis, and E.~Learned-Miller, ``Multi-view convolutional neural networks for 3d shape recognition,'' in \emph{Proceedings of the IEEE international conference on computer vision}, 2015, pp. 945--953.

\bibitem{maturana2015voxnet}
D.~Maturana and S.~Scherer, ``Voxnet: A 3d convolutional neural network for real-time object recognition,'' in \emph{2015 IEEE/RSJ international conference on intelligent robots and systems (IROS)}.\hskip 1em plus 0.5em minus 0.4em\relax IEEE, 2015, pp. 922--928.

\bibitem{li2018pointcnn}
Y.~Li, R.~Bu, M.~Sun, W.~Wu, X.~Di, and B.~Chen, ``Pointcnn: Convolution on x-transformed points,'' \emph{Advances in neural information processing systems}, vol.~31, 2018.

\bibitem{thomas2019kpconv}
H.~Thomas, C.~R. Qi, J.-E. Deschaud, B.~Marcotegui, F.~Goulette, and L.~J. Guibas, ``Kpconv: Flexible and deformable convolution for point clouds,'' in \emph{Proceedings of the IEEE/CVF international conference on computer vision}, 2019, pp. 6411--6420.

\bibitem{peyghambarzadeh2020point}
S.~M. Peyghambarzadeh, F.~Azizmalayeri, H.~Khotanlou, and A.~Salarpour, ``Point-planenet: Plane kernel based convolutional neural network for point clouds analysis,'' \emph{Digital Signal Processing}, vol.~98, p. 102633, 2020.

\bibitem{lu2020pointngcnn}
Q.~Lu, C.~Chen, W.~Xie, and Y.~Luo, ``Pointngcnn: Deep convolutional networks on 3d point clouds with neighborhood graph filters,'' \emph{Computers \& Graphics}, vol.~86, pp. 42--51, 2020.

\bibitem{wang2019dynamic}
Y.~Wang, Y.~Sun, Z.~Liu, S.~E. Sarma, M.~M. Bronstein, and J.~M. Solomon, ``Dynamic graph cnn for learning on point clouds,'' \emph{ACM Transactions on Graphics (tog)}, vol.~38, no.~5, pp. 1--12, 2019.

\bibitem{yang2019modeling}
J.~Yang, Q.~Zhang, B.~Ni, L.~Li, J.~Liu, M.~Zhou, and Q.~Tian, ``Modeling point clouds with self-attention and gumbel subset sampling,'' in \emph{Proceedings of the IEEE/CVF conference on computer vision and pattern recognition}, 2019, pp. 3323--3332.

\bibitem{engel2021point}
N.~Engel, V.~Belagiannis, and K.~Dietmayer, ``Point transformer,'' \emph{IEEE access}, vol.~9, pp. 134\,826--134\,840, 2021.

\bibitem{vaswani2017attention}
A.~Vaswani, ``Attention is all you need,'' \emph{Advances in Neural Information Processing Systems}, 2017.

\bibitem{qiu2021geometric}
S.~Qiu, S.~Anwar, and N.~Barnes, ``Geometric back-projection network for point cloud classification,'' \emph{IEEE Transactions on Multimedia}, vol.~24, pp. 1943--1955, 2021.

\bibitem{xiang2021walk}
T.~Xiang, C.~Zhang, Y.~Song, J.~Yu, and W.~Cai, ``Walk in the cloud: Learning curves for point clouds shape analysis,'' in \emph{Proceedings of the IEEE/CVF international conference on computer vision}, 2021, pp. 915--924.

\end{thebibliography}
}

\end{document}